\documentclass[journal]{IEEEtran}
\IEEEoverridecommandlockouts
\usepackage{cite}
\usepackage{amsmath,amssymb,amsfonts}
\usepackage{algorithm}
\usepackage[noend]{algpseudocode}
\usepackage{graphicx}
\usepackage{textcomp}
\usepackage{xcolor}
\def\BibTeX{{\rm B\kern-.05em{\sc i\kern-.025em b}\kern-.08em
    T\kern-.1667em\lower.7ex\hbox{E}\kern-.125emX}}
\usepackage[colorlinks,urlcolor=blue,linkcolor=blue,citecolor=blue]{hyperref}
\usepackage{soul}
\usepackage{float}
\usepackage{bm}
\usepackage{booktabs} 
\usepackage{array} 
\usepackage{multirow}
\sethlcolor{yellow}

\begin{document}
\title{Exploring Energy Landscapes for Minimal Counterfactual Explanations:  Applications in Cybersecurity and Beyond}
\author{Spyridon Evangelatos\thanks{S.~Evangelatos and E.~Veroni are with the Research \& Innovation Development Department, Netcompany-Intrasoft S.A.,  Luxembourg and the Department of Electronic Engineering, Hellenic Mediterranean University, Greece, (emails: sevangelatos@netcompany.com and everoni@netcompany.com).}, Eleni Veroni, Vasilis Efthymiou\thanks{V.~Efthymiou and G.~Papadopoulos are with Department of Informatics and Telematics, Harokopio University of Athens, Greece, (emails: vefthym@hua.gr and g.th.papadopoulos@hua.gr).}, Christos Nikolopoulos, \IEEEmembership{Member, IEEE}\thanks{C.~Nikolopoulos is with the Department of Electronic Engineering, Hellenic Mediterranean University, Greece, (email: cnikolo@hmu.gr).}, Georgios Th. Papadopoulos, \IEEEmembership{Member, IEEE} and Panagiotis Sarigiannidis, \IEEEmembership{Member, IEEE}\thanks{P.~Sarigiannidis is with the Department of Electrical and Computer Engineering, University of Western Macedonia, Greece, (email: psarigiannidis@uowm.gr).}}


\maketitle

\begin{abstract}
Counterfactual explanations have emerged as a prominent method in Explainable Artificial Intelligence (XAI), providing intuitive and actionable insights into Machine Learning model decisions. In contrast to other traditional feature attribution methods that assess the importance of input variables, counterfactual explanations focus on identifying the minimal changes required to alter a model's prediction, offering a ``what-if'' analysis that is close to human reasoning. In the context of XAI, counterfactuals enhance transparency, trustworthiness and fairness, offering explanations that are not just interpretable but directly applicable in the decision-making processes.
 
In this paper, we present a novel framework that integrates perturbation theory and statistical mechanics to generate minimal counterfactual explanations in explainable AI. We employ a local Taylor expansion of a Machine Learning model's predictive function and reformulate the counterfactual search as an energy minimization problem over a complex landscape. In sequence, we model the probability of candidate perturbations leveraging the Boltzmann distribution and 
use simulated annealing for iterative refinement. Our approach systematically identifies the smallest modifications required to change a model's prediction while maintaining plausibility. Experimental results on benchmark datasets for cybersecurity in Internet of Things environments, demonstrate that our method provides actionable, interpretable counterfactuals and offers deeper insights into model sensitivity and decision boundaries in high-dimensional spaces.
\end{abstract}


\section{Introduction}
\IEEEPARstart{A}{rtificial Intelligence} (AI) systems are increasingly deployed in critical decision-making processes such as healthcare, finance, manufacturing and law enforcement. While these systems often achieve state-of-the-art performance, the complexity of their underlying models can render them opaque to end users and stakeholders. XAI has emerged as one of the most eminent areas of research striving to bring transparency and build trust by explaining the rationale behind AI decisions~\cite{Ronge:2024}. Among the different approaches within XAI, counterfactual explanations have gained particular prominence since they specify the minimal changes to an input needed to alter a model’s output, thereby offering actionable insights to those affected by automated decisions. In contrast with causal explanations, which focus on identifying the underlying factors that led to a decision~\cite{Celar:2023}, counterfactuals emphasize how a different outcome could have been achieved, enhancing the transparency and trustworthiness of AI decisions.

Counterfactual explanation is an example-based post-hoc explanation method used in XAI to provide insights into how Machine Learning models make predictions~\cite{Stepin:2021}. They describe hypothetical scenarios in order to identify the minimal modifications to an input $\bm{x}\in\mathbb{R}^{n}$ that can alter the prediction of the model $f$, thus offering invaluable insights into the model's decision-making process. This approach highlights the most influential features of $\bm{x}$ that drive the model's decision while also delineates the path necessary to traverse the decision boundary with minimal deviation from the original input.

Traditional feature attribution methods, which assign importance scores to input features, often fail to provide actionable insights, as they do not reveal how a model's prediction could have been different under a slightly altered input~\cite{Mothilal:2021FeatAttrib}. Counterfactual explanations address this gap by specifying the minimal modifications required to achieve a different model outcome, offering a form of explanation that aligns with human reasoning and decision-making. The challenge in generating counterfactuals lies in the complexity of navigating the model’s decision boundary while ensuring that the proposed perturbations remain minimal and feasible. Many existing counterfactual generation methods, suffer from critical limitations, including sensitivity to local minima, instability across different runs, and an inability to properly balance exploration and exploitation. 

Perturbation theory is a versatile mathematical framework for approximating complex systems~\cite{Sacchetti:2009}. Its methods are used to analyze how small deviations from a known state affect the overall behavior of a system. When combined with counterfactual explanations, it enables the systematic identification of minimal input modifications that can change a model's output. Statistical mechanics employs probabilistic methods and energy minimization principles in order to bridge microscopic interactions with macroscopic phenomena~\cite{Huang:1987}. These principles can be applied to frame the search for minimal input changes as an energy minimization problem, where low-energy states correspond to plausible modifications that shift a model's output. 

This paper introduces a novel framework that generates minimal, actionable, counterfactual explanations in the field of explainable AI. Our framework provides a rigorous way to explore model sensitivities, escape local minima and deliver robust, interpretable solutions that foster deeper trust and transparency in AI-driven decision-making processes. It addresses the aforementioned limitations of existing counterfactual generation methods proposing an energy-based framework inspired by statistical mechanics, where the counterfactual search process is formulated as an energy minimization problem. The combination of perturbation theory and simulated annealing enables the proposed framework to systematically explore the decision space in order to identify optimal perturbations that shift the model’s prediction while preserving plausibility. Moreover, the introduction of an entropy term allows for a controlled exploration-exploitation trade-off, preventing the counterfactual search from being trapped in local optima.

The rest of the paper is organized as follows. Section~\ref{Sec:RelWork} provides an overview of the relevant work on counterfactual explanations, perturbation theory and the application of statistical mechanics to XAI. Section~\ref{Sec:FraDes} describes the framework's foundations and principles, followed by a detailed algorithmic implementation. In sequence, we validate our approach through extensive simulations described in detail in Section~\ref{Sec:ResAna} and discuss the implications of our results. For completence purposes, Section~\ref{Sec:CompOther} presents a detailed coparison of our proposed frameowrk with other state-of-the-art techniques. Section~\ref{SecCompComplex} analyses the computational complexity of the Free-Energy based algorithms whereas Section~\ref{Sec:Robust} provides the evaluation of the stability of our proposed framework under small and adversarial perturbations. Last, Section~\ref{Sec:ConFut} concludes this paper with a summary of our findings and future directions for research.

\section{Related Work}\label{Sec:RelWork}
Counterfactual explanations have emerged as a prominent approach in XAI~\cite{Rawal:2022}. They build upon the seminal work of~\cite{Wachter:2017} where counterfactual explanations are framed as actionable modifications to input data that yield different predictions without requiring full model transparency. The authors provide an algorithmic process which uses a distance metric to quantify how much an input must change, and then search for a new point that is both feasible and yield the desired prediction. The proposed method is model-agnostic since it was focusing only on inputs and outputs making it applicable to a wide variety of Machine Learning algorithms without requiring access to model parameters or architecture. Nevertheless, the framework did not consider real-world constraints, nor did it address how counterfactual methods might detect or mitigate systemic biases that appear at the group level, an increasingly important concern in AI ethics~\cite{Huang:2023}. Last, computational challenges or optimizations required when dealing with very high-dimensional data and complex models at scale, were not explored.

Building on this concept, several subsequent studies explored diverse methods to generate counterfactuals, ranging from mixed-integer linear optimization techniques~\cite{Verma:2024} to more complex, model-agnostic approaches~\cite{Guidotti:2024}. In their ground-breaking study, the authors of ~\cite{Mothilal:2020} presented the DiCE framework for generating diverse counterfactual explanations, leveraging the Determinantal Point Processes. They introduced metrics that evaluate the feasibility of the suggested changes within the user's context, ensuring that the recommended actions are realistic and attainable. However, they do not explicitly account for the stability and robustness of counterfactual explanations under adversarial perturbations or model uncertainty, making it susceptible to generating counterfactuals that are highly sensitive to minor input variations~\cite{Bayrak:2024}. 

Other notable examples include methods tailored to linear models for actionable recourse~\cite{Ustun:2019}, addressing the ability of individuals to alter actionable input variables of a Machine Learning models, to change an unfavorable model prediction into a favorable one. The authors employed integer programming (IP) techniques in order to identify minimal and feasible changes that individuals can make to achieve a desired prediction, further considering constraints such as feature immutability and actionability. Nevertheless, the proposed framework is limited to linear models, making it unsuitable for capturing the complex, nonlinear decision boundaries of modern Machine Learning models~\cite{Delaunay:2022}.

Other methods have focused on exploring the space of possible counterfactuals by formulating the search as an optimization or heuristic-driven problem, allowing them to generate multiple, coherent explanations that satisfy the model's decision boundary constraints~\cite{Russell:2019}. This approach does not require access to the internal parameters or gradients of a model, making them applicable to ``black-box'' Machine Learning systems. Despite its ability to generate diverse counterfactuals, the search-based strategies often lack theoretical guarantees on optimality and stability, leading to inconsistent results across different runs and potentially suboptimal counterfactuals that do not minimize perturbation effectively. 

Recent novel approaches~\cite{Karimi:2020}, formulate the problem of generating counterfactual explanations as a sequence of satisfiability problems, where both the predictive model and the distance function are represented as logic formulas. This method is notably model-agnostic, accommodating various model types, including non-linear, non-differentiable and non-convex models, and is also agnostic to data types and distance metrics, effectively handling heterogeneous features and multiple norms. However, this approach relies heavily on formal verification tools~\cite{Leofante:2023}, which can lead to scalability issues when applied to high-dimensional feature spaces or large-scale datasets, making it computationally expensive in practical deployments. Additionally, while it ensures logical consistency, the generated counterfactuals may lack smoothness and flexibility compared to methods that incorporate gradient-based refinements or energy-based exploration.  

These methods emphasize the importance of minimality and plausibility, ensuring that the proposed counterfactuals are both interpretable and achievable in practice. Nowadays, counterfactual explanations have evolved from simpler case-based reasoning approaches to more sophisticated optimization-driven techniques that provide explicit paths for altering model outcomes~\cite{Ferrario:2022}.  

In parallel, perturbation theory~\cite{Kato:1966} has been extensively used in order to analyze the impact of small changes and approximate complex systems such as Deep Neural Networks~\cite{Ivanovs:2021}. In the context of Machine Learning, local Taylor expansions provide a powerful tool for sensitivity analysis, as seen in various works such as the Local Interpretable Model-Agnostic Explanations (LIME) method~\cite{Ribeiro:2016} which learns an interpretable, local surrogate model around each prediction by sampling perturbed inputs and observing how the black-box classifier behaves, thus explaining individual predictions without requiring access to model internals. However, while these techniques offer valuable insights into local behavior, they often struggle with non-linearities and high-dimensional spaces, motivating the search for more robust optimization frameworks~\cite{Yang:2023}.

Statistical mechanics has been employed in various fields, from computational biology and quantitative fianance to Internet-of-Things (IoT) networks~\cite{Evangelatos:2023} and error-correcting codes~\cite{Sourlas:1994}. Recently, they have inspired novel approaches to AI, particularly in the realm of energy-based models and optimization~\cite{Decelle:2023}. Concepts such as the Boltzmann distribution and simulated annealing have been used to recast learning problems as energy minimization tasks, enabling a global perspective on model behavior~\cite{Krzakala:2024}. Although this perspective has found applications in generative modeling and deep learning, its integration with counterfactual explanations remains relatively unexplored.

Our work bridges these research threads by combining perturbation theory and statistical mechanics to generate minimal counterfactual explanations. This integration leverages local approximations of model sensitivity and adopts an energy-based framework to navigate complex, high-dimensional landscapes. The result is a robust and interpretable method that addresses the limitations of existing counterfactual approaches and offers a new pathway for transparent AI decision-making.

\begin{algorithm}[t]
\caption{Iterative Refinement via Simulated Annealing}\label{alg:cf}
\begin{algorithmic}[1]
\State \textbf{Input:} Original instance $\bm{x}$, target threshold $c$, learning rate $\alpha$, inverse temperature $\beta$, tolerance $\epsilon$, hyperparameters $\lambda$ and $\mu$, regularization on perturbation size $C_{1}$, decision boundary sensitivity term $C_{2}$.
\State \textbf{Initialize:} Set $\bm{x}^{(0)} \gets \bm{x}$ 
\State Compute the initial free energy:
\[
\mathcal{F}_{\beta}(\bm{x}^{(0)}) = \mathcal{E}(\bm{x}^{(0)}) - \frac{1}{\beta} \mathcal{S}_{\beta}(\bm{x}^{(0)}),
\]
where $\mathcal{E}(\bm{x}) = \|\bm{x} - \bm{x}^{(0)}\|_{\ell_{2}} + \lambda R(\bm{x}) + \mu\, \bigl|f(\bm{x})- c\bigr|$ and $\mathcal{S}_{\beta}(\bm{x}) = - \int p(\Delta \bm{x}) \ln \Big[p(\Delta \bm{x})\Big] \, d(\Delta \bm{x})$.
\For{$t = 0, 1, 2, \ldots$}
    \State \textbf{Local Approximation:} Compute the Taylor expansion of $f$ around $\bm{x}^{(t)}$ to obtain $\nabla f(\bm{x}^{(t)})$ (and, if available, i.e., the ML model provides access to it, $H(\bm{x}^{(t)})$)
    \State \textbf{Free Energy Evaluation:} Compute 
    \[
    \mathcal{F}_{\beta}(\bm{x}^{(t)}) = \mathcal{E}(\bm{x}^{(t)}) - \frac{1}{\beta} \mathcal{S}_{\beta}(\bm{x}^{(t)}).
    \]
    \State \textbf{Gradient Update:} Compute the candidate update
    \[
    \tilde{\bm{x}} \gets \bm{x}^{(t)} - \alpha\, \nabla_{\bm{x}} \mathcal{F}_{\beta}(\bm{x}^{(t)}).
    \]
    \State Compute free energy difference: 
    \[
    \Delta \mathcal{F}_{\beta} \gets \mathcal{F}_{\beta}(\tilde{\bm{x}}) - \mathcal{F}_{\beta}(\bm{x}^{(t)})
    \]
    \State \textbf{Gradient Sensitivity Check:} 
    \If{$\|\nabla_{\bm{x}} \mathcal{F}_{\beta}(\tilde{\bm{x}})\|_{\ell_{2}} > C_1$}
        \State $\alpha \gets \frac{1}{2}\alpha$, Reduce learning rate
        \State Recompute $\tilde{\bm{x}}$ with adjusted $\alpha$
    \EndIf
    \State \textbf{Hessian Stability Check (if available):}
    \If{$\lambda_{\max} (H_{\mathcal{F}}(\bm{x}^{(t)})) > C_2$}
        \State Reject update and revert to $\bm{x}^{(t)}$
    \EndIf
    \State \textbf{Simulated Annealing Acceptance:} 
    \If{$\Delta \mathcal{F}_{\beta} \le 0$ \textbf{or} $\exp(-\beta\, \Delta \mathcal{F}_{\beta}) > \mathrm{rand}(0,1)$}
        \State $\bm{x}^{(t+1)} \gets \tilde{\bm{x}}$
    \Else
        \State $\bm{x}^{(t+1)} \gets \bm{x}^{(t)}$
    \EndIf
    \State \textbf{Convergence Check:} 
    \If{$\lvert f(\bm{x}^{(t+1)}) - c\rvert < \epsilon$ }
        \State \textbf{Adversarial Robustness Check:} Perturb final $\bm{x}_{cf}$ with $\|\bm{\eta}\| \leq \xi$ and verify:
        \[
        \mathcal{F}_{\beta}(\bm{x}_{cf} + \bm{\eta}) \approx \mathcal{F}_{\beta}(\bm{x}_{cf})
        \]
        \If{counterfactual fails robustness check}
            \State Repeat optimization with updated constraints.
        \EndIf
        \State \Return $\bm{x}^{(t+1)}$
    \EndIf
\EndFor
\end{algorithmic}
\end{algorithm}

\section{Framework Description}\label{Sec:FraDes}
Let us denote by $f: \mathbb{R}^n \to \mathbb{R}$ a predictive model trained on a dataset $\mathrm{D}$. Then, given an input instance $\bm{x}\in \mathbb{R}^n$ with model output $f(\bm{x})$, our goal is to find a counterfactual $\bm{\hat{x}}$ \emph{close} to $\bm{x}$, such that $f(\bm{\hat{x}})$ differs from $f(\bm{x})$, i.e., 
\begin{equation}\label{eq:cf_baseline}
  \min_{\bm{\hat{x}}} \; \|\bm{x} - \bm{\hat{x}}\|_{\ell_{2}}
  \quad \text{subject to} \quad f(\bm{\hat{x}}) \neq f(\bm{x}),
\end{equation}
where $\|\bm{x} - \bm{\hat{x}}\|_{\ell_{2}}$ is the Euclidean distance between the original input $\bm{x}$ and its counterfactual $\bm{\hat{x}}$ that captures the overall magnitude of change required to alter the model’s prediction. The choice of the $\ell_{2}$-norm has been made in order to encourage small (often smoother) perturbations across all dimensions. Nevertheless, in case we want to accommodate more sparse changes, e.g., altering the fewest possible features to change the model's prediction, we can opt for the $\ell_{1}$-norm.

Solving directly the counterfactual optimization problem defined in Eq.~\eqref{eq:cf_baseline} can be extremelly challenging - if not infeasible - for complex or non-linear models. For this purpose, we apply perturbation theory in order to obtain a tractable local approximation and translate the high-level objective into a simpler optimization problem in the vicinity of the original input.

To handle complex, non-linear models, we use a local Taylor expansion to approximate $f$ around $\bm{x}$:
\begin{eqnarray}\label{eqn:taylor}
f(\bm{x} + \Delta \bm{x}) & \approx & f(\bm{x}) + \nabla f(\bm{x})^\top \Delta \bm{x} \nonumber\\ 
&+& \frac{1}{2} \Delta \bm{x}^\top H(\bm{x}) \Delta \bm{x} + \mathcal{O}(\|\Delta \bm{x}\|^3),
\end{eqnarray}
where $\Delta \bm{x} = \bm{\hat{x}} - \bm{x}$,  $\nabla f(\bm{x})$ is the gradient at $\bm{x}$ and $H(\bm{x})$ is the Hessian encoding the second-order partial derivatives of $f$ at $\bm{x}$. This expansion captures the model’s local behavior and helps us formulate a tractable \emph{local} optimization problem for identifying $\Delta \bm{x}$ that changes $f$ sufficiently while staying as close as possible to the original input $\bm{x}$.

It is worth highlighting that Eq.~\eqref{eqn:taylor} include the term $\mathcal{O}(\|\Delta \bm{x}\|^3)$ representing the higher-order terms in the Taylor expansion of the model's decision function $f(\bm{x})$ around $\bm{x}$. While the quadratic approximation captures local curvature through the Hessian, the cubic term represents nonlinear corrections that become significant in highly non-convex landscapes. Although these higher-order terms are typically neglected for computational efficiency, their presence quantifies the approximation error introduced by truncating the expansion at the second-order term. In practice, the quadratic approximation is often sufficient, but for models with highly nonlinear structures, incorporating higher-order corrections could enhance the accuracy of counterfactual search methods.

Perturbation theory alone provides a local approximation of how the model’s output changes near a given point, but it does not inherently address how to globally navigate the space of possible perturbations or escape local minima~\cite{Jin:2017:saddle}. Thus, we recast the search for a feasible, minimal modification as an energy minimization problem under a statistical mechanics framework, combining local expansions with a broader, energy-focused perspective, resulting in more robust and realistic counterfactual generation.  

We define an effective energy function:
\begin{equation}\label{eqn:energy}
\mathcal{E}(\Delta \bm{x}) = \|\Delta \bm{x}\|_{\ell_{2}} + \lambda R(\Delta \bm{x}) + \mu \big| f(\bm{x}+\Delta \bm{x}) - c \big|,
\end{equation}
where $R(\Delta x)$ is a regularization term that enforces domain-appropriate modifications (e.g., sparsity, bounded changes, or other feasibility constraints), ensuring the resulting counterfactual remains realistic, $\lambda$ and $\mu$ are hyperparameters balancing the trade-off between minimizing the perturbation and achieving the desired change in the model's prediction and $c$ is the target output  indicating the threshold or boundary to be crossed by the model.

With the energy function defined in Eq.~\eqref{eqn:energy}, we now reinterpret the counterfactual search as a probabilistic process by applying the Boltzmann distribution, which assigns higher probabilities to lower-energy states according to the corrsponding Boltzmann (Gibbs) distribution:
\begin{equation}\label{eqn:boltzmann}
p(\Delta \bm{x}) = \frac{1}{\mathcal{Z}} \exp\Bigl(-\beta \mathcal{E}(\Delta \bm{x})\Bigr),
\end{equation}
with the partition function that ensures normalization being
\begin{equation}
\mathcal{Z}(\beta) = \int \exp\Bigl(-\beta \mathcal{E}(\Delta \bm{x})\Bigr) \mathrm{d}\Delta \bm{x},
\end{equation}
and $\beta$ representing the inverse temperature that controls how strongly we penalize higher-energy states. Through this formulation minimal and feasible perturbations are prioritized while efficient exploration of the energy landscape can be performed using sampling methods such as simulated annealing.

The Boltzmann distribution assigns a probability to each perturbation based on its energy, and this naturally leads to a free energy formulation that balances 
the energy $\mathcal{E}(\Delta \bm{x})$ with an entropy term in order to capture both the cost of a perturbation and the density of neighboring low-energy states. Therefore, we define the free energy function
\begin{equation}\label{eqn:FreeEnergy}
\mathcal{F}_{\beta}(\Delta \bm{x}) = \mathcal{E}(\Delta \bm{x}) - \frac{1}{\beta}\mathcal{S}_{\beta}
\end{equation}
where $\mathcal{S}_{\beta}$ represents the entropy of the perturbation distribution, defined as 
\begin{equation}\label{eqn:EntropyTerm}
\mathcal{S}_{\beta} = -\int p(\Delta \bm{x})\ln \Big[p(\Delta \bm{x})\Big]\mathrm{d}\Delta \bm{x}
\end{equation}

Based on the free energy function in Eq.~\eqref{eqn:FreeEnergy}, we treat the counterfactual search as a minimization problem 
\begin{equation}
  \min_{\Delta\bm{\hat{x}}} \; \mathcal{F}_{\beta}(\Delta \bm{x}) 
  \quad \text{subject to} \quad f(\bm{\hat{x}}+\Delta\bm{x}) \neq f(\bm{x}),
\end{equation}
Minimizing $\mathcal{F}_{\beta}(\Delta \bm{x})$ ensures both minimal perturbation and a robust, diverse set of counterfactuals. Iterative refinement via simulated annealing achieves this by gradually increasing the inverse temperature $\beta$, which transitions the search from broad exploration of the energy landscape to focused convergence on the optimal, low-energy counterfactual solution. This process is described in Algorithm~\ref{alg:cf}.

The energy landscape induced by Eq.~\eqref{eqn:FreeEnergy} offers a rich pespective on the model’s sensitivity. In this framework, each potential perturbation $\Delta\bm{x}$ corresponds to a point on a high-dimensional landscape where low-energy regions represent small, feasible modifications that can change the model's prediction. A cluster of low-energy solutions implies that the modification is not a singular, fragile solution but part of a broader basin where the model's output is likely to change in a reliable manner. Such interpretation assist us in understanding which features are critical and how small shifts can yield significant changes in the model's behavior. 

Moreover, examining the free energy landscape allows for a deeper analysis of the model's  behavior near the decision boundaries. The gradient of the landscape, derived from the local Taylor expansion, indicates how sensitive the model is to perturbations in specific directions, informing us on  the relative importance of different features. Th hyperparameters $\lambda$, $\mu$ and $\beta$ can further control the trade-off between minimality and robustness, offering an interpretable means to adjust the explanation to suit different application contexts.

Ensuring that counterfactual explanations are actionable means that the modifications recommended by the model must be both feasible and practical in real-world scenarios~\cite{Wang:2021}. In our framework, actionability is addressed by incorporating domain-specific constraints into the energy function through the regularization term $R(\Delta\bm{x})$ (controlled for example by the parameters $\lambda$ in Eq.~\eqref{eqn:energy}). For example, features such as age, gender or other immutable attributes can be constrained by assigning them high or even infinite cost when modified, effectively preventing unrealistic counterfactuals. 

Furthermore, the free energy formulation - by integrating both energy and entropy - ensures that the counterfactual is derived from a set of plausible and robust modifications. The entropy term $\mathcal{S}_{\beta}(\Delta\bm{x})$ encourages the discovery of solutions that are supported by a diverse set of nearby perturbations. This diversity is crucial for actionability, as it implies that the recommendation is not an isolated outlier but part of a coherent region of the input space where similar, feasible changes yield the desired outcome.

\begin{figure}[t]
\includegraphics[width=0.48\textwidth]{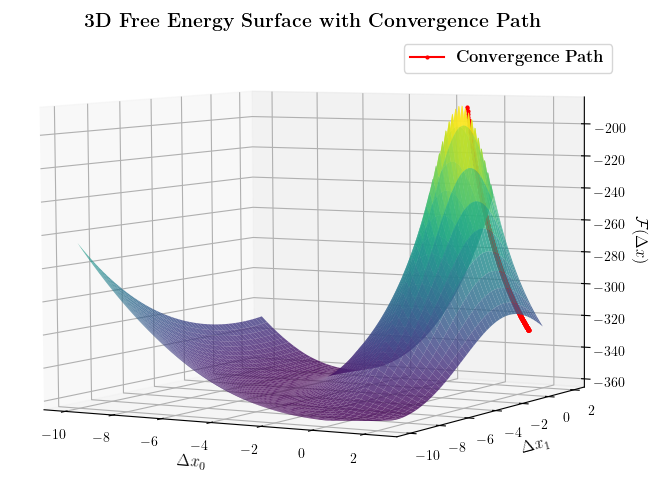}
\caption{A 3D surface plot of the free energy function $\mathcal{F}_{\beta}(\Delta\bm{x})$ over the perturbation space. The parameters used are $\Delta \bm{x} \in [-10,2]^2$, $\lambda=1.0$, $\mu=1.0$, $w_{i}\sim\mathcal{U}(1,10)$, $\bm{x}_0=[3,3]$, $c=0$, $\alpha=0.01$, $\beta=0.01$, $n=500$ iterations, $\epsilon = 10^{-4}$.}\label{fig:EnergyLandscape}
\end{figure}

\section{Results and Analysis}\label{Sec:ResAna}
Let us consider an  IoT network deployed in a critical infrastructure environment where numerous sensors continuously monitor wireless signal strengths, network traffic metrics, and various security vulnerability indicators. In this setting, the cybersecurity system relies on a Machine Learning classifier $f$ to assess the current state of the network and detect potential intrusions or anomalies. Our goal is to generate counterfactual explanations that reveal the minimal modifications to the sensor readings or network metrics necessary to alter the classifier’s decision. Such counterfactuals serve two key purposes; they assist us understanding which features are most influential in triggering a security alert, and they provide actionable insights into how minor adjustments could potentially prevent false alarms or mitigate emerging threats. In such a scenario, a candidate energy function that reflects both the magnitude of the perturbation and the criticality of changes to key security-related features 
can be defined as
\begin{equation}\label{eqn:EnergyFunction}
\mathcal{E}(\Delta \bm{x}) = \|\Delta \bm{x}\|_{\ell_{2}} + \lambda \sum\limits_{i\in\mathrm{V}}w_{i}\big|\Delta x_{i}\big| + \mu \big| f(\bm{x}+\Delta \bm{x}) - c \big|,
\end{equation}
where $\mathrm{V}$ denotes the set of security-critical features, and $w_{i}$ the weights that quantify the risk or importance of modifying each of these features, set either by cybersecurity experts or potentially by automated algorithms performing feature importance scores (e.g.~\cite{Lundberg:2017:SHAP}). This formulation penalizes changes to features that are critical to the system's security, thereby aligning the counterfactual explanation with real-world constraints in IoT cybersecurity contexts. 

\begin{figure}[t]
\includegraphics[width=0.5\textwidth]{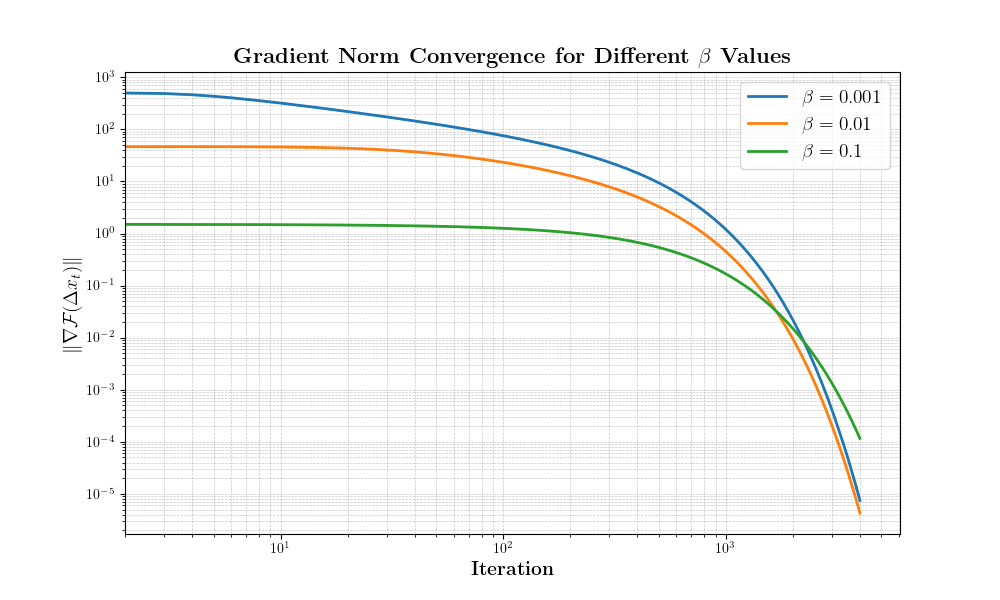}
\caption{Gradient norm convergence for different $\beta$ values. The parameters used were $\Delta \bm{x} \in [-10,2]^2$, $\lambda=1.0$, $\mu=1.0$, $w_{i}\sim\mathcal{U}(1,10)$, $\bm{x}_0=[3,3]$, $c=0$, $\alpha=10^{-3}$, $n=6000$ iterations, $\epsilon = 10^{-4}$.}\label{fig:NormConver}
\end{figure}

The input features may include a variety of sensor and network metrics. For instance, wireless sensor readings might involve temperature, humidity, or vibration levels, while network traffic metrics could cover packet loss, latency, or throughput. Additionally, security vulnerability indicators such as the frequency of unauthorized access attempts or port scan counts may also be included. Performing Algorithm~\ref{alg:cf} in the input features of the IoT network, we allow small modifications to non-critical features (or only slight adjustments to critical ones, which are heavily penalized) so that the system's decision can be altered with minimal, realistic changes.

The 3D surface in Figure~\ref{fig:EnergyLandscape} represents the free energy $\mathcal{F}(\Delta \bm{x})$ as a function of the two perturbation coordinates $\Delta x_{0}$ and $\Delta x_{1}$. Regions with higher surface elevations indicate higher free-energy values, while the lower ``valleys'' correspond to more desirable points with minimal free energy. The color gradient provides a quick visual guide to where the algorithm is more or less likely to converge if it seeks to minimize $\mathcal{F}$. From left to right, one can see that there is a substantial ``hill'' on the right side of the plot, which the algorithm avoids or traverse carefully, whereas the left region of the landscape sits at a lower free-energy level.

\begin{figure*}[!t]
\includegraphics[width=\textwidth]{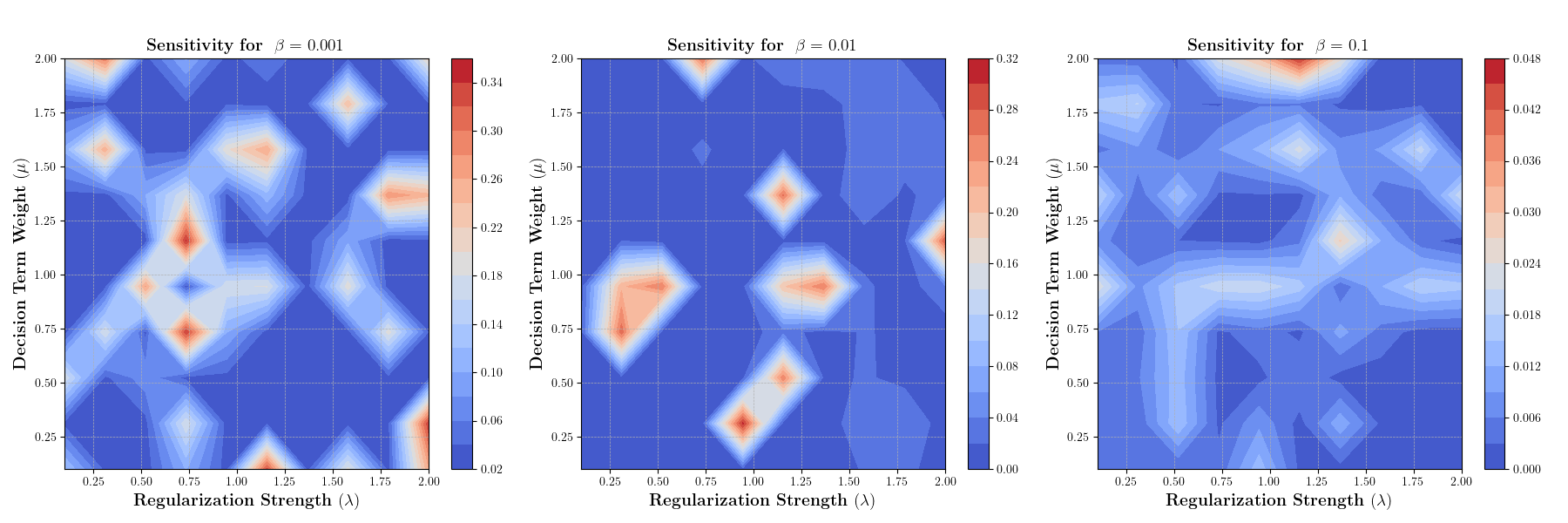}
\caption{Contour plots illustrating the sensitivity of counterfactual stability to variations in the regularization strength, $\lambda$, and the decision term weight $\mu$ for different values of the inverse temperature $\beta$. Higher stability regions (red) indicate optimal parameter pairs that yield more robust counterfactuals, while lower stability regions (blue) highlight configurations that lead to higher sensitivity to perturbations.}\label{fig:HyperparametersSensitivity}
\end{figure*}

Overlaid on this surface is a red trajectory indicating the convergence path taken by our algorithm starting from an initial point and iteratively refining its solution. The path's progression in the plot shows that the algorithm began in a higher-energy region and then descended toward a lower-energy valley. The red markers indicate each iteration's position, highlighting where the algorithm ``stepped'' to find improvements in the free-energy value. Notably, the final segment of the path is in a region of relatively lower elevation, suggesting that our method has successfully identified a minimum or near-minimum in the free-energy landscape.

Figure~\ref{fig:NormConver} illustrates the gradient norm $\|\nabla\mathcal{F}(\Delta \bm{x}^{(t)})\|$ at each iteration for different values of the inverse temperature $\beta$. As the algorithm refines its solution $\Delta \bm{x}^{(t)}$, the decreasing gradient norm indicates that the algorithm is converging to a stationary point, whether local or global. However, because the gradient can vanish at local minima, saddle points, or even plateaus, a near-zero gradient norm alone does not guarantee that the solution is globally optimal. Rather, it confirms that local improvements have diminished, suggesting that the method has reached (or is very close to) a stationary point in the energy landscape.

In this simulation, the three $\beta$ values $\{0.001, 0.01, 0.1\}$ were chosen to explore distinct regimes of the acceptance probability and illustrate how the algorithm transitions from a relatively “exploratory” phase to a more “exploitative” one. In simulated annealing or related methods, 
 $\beta$ controls how readily uphill moves are accepted. With very small  $\beta$ the algorithm has a relatively high chance of accepting uphill steps, promoting broader exploration of the free-energy landscape at the cost of slower, more meandering convergence. Conversely, as  $\beta$ increases (e.g., to 
$0.1$), the algorithm becomes more selective in accepting moves that do not reduce the free energy, thus favoring quicker descent into local minima.

Moreover, these $\beta$ values are neither so large as to cause the algorithm to get stuck immediately in local minima, which can happen if $\beta$ is extremely high, nor so small that it behaves nearly randomly. This balance allows us to demonstrate how moderate $\beta$ values produce consistent decreases in the gradient norm while still permitting enough stochasticity to avoid trivial local minima. Additionally, empirical testing revealed that $\beta$ values in the range of $1$ to $10$ caused the algorithm to oscillate without consistently reducing the gradient norm. In that regime, the acceptance of increases in free energy is overly constrained, limiting meaningful exploration. At the same time, it is not low enough to encourage sufficient reductions for convergence. As a result, the algorithm effectively remains in a quasi-random walk, preventing it from settling into a stable minimum. Hence, although $\beta$ values between $1$ and $10$ might theoretically balance exploration and exploitation in some scenarios, in this particular setup they fail to reduce the gradient norm consistently, leading to non-convergence in practice.
 
From a different perspective, when the inverse temperature parameter $\beta$ approaches zero, the entropy term in the Free-Energy function defined in Eq.~\eqref{eqn:FreeEnergy} influences signifacantly the counterfactual generation process. Thus $\beta$ acts as a parameter that controls the trade-off between energy minimization and entropy maximization.

\begin{figure*}[t]
\includegraphics[width=\textwidth]{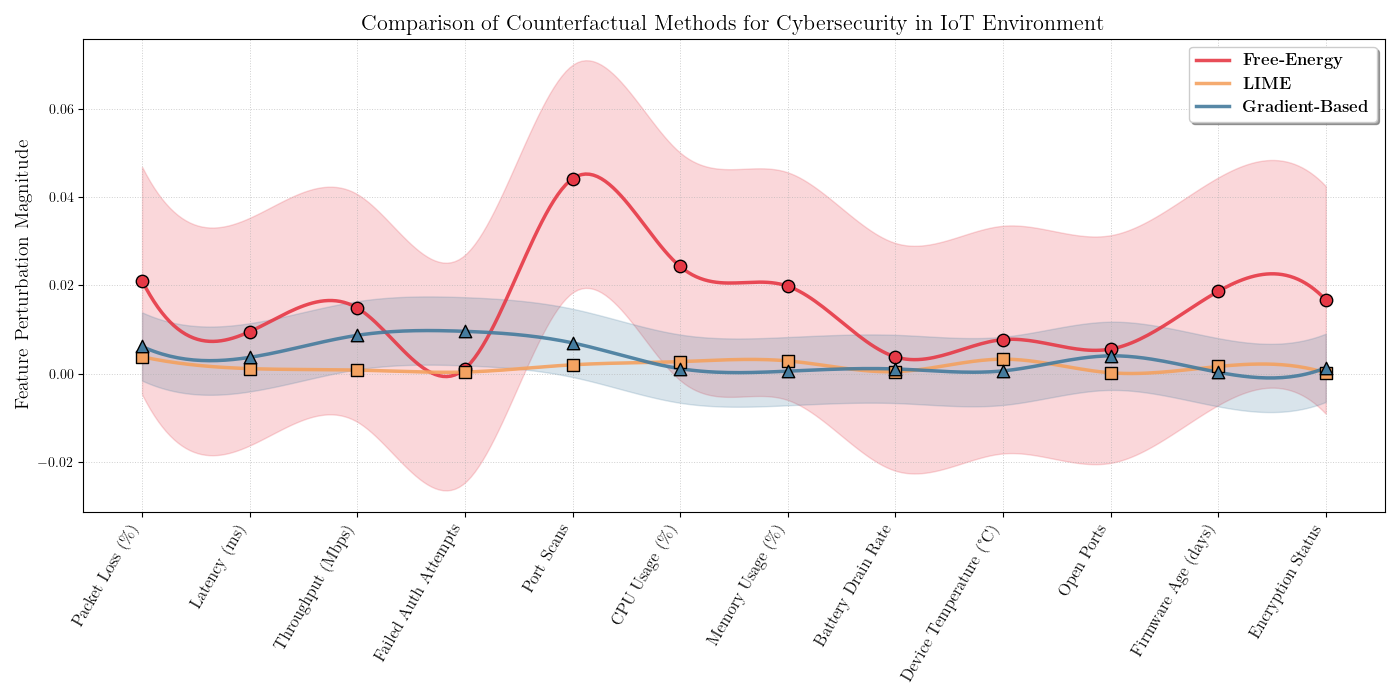}
\caption{Comparison of three counterfactual methods for cybersecurity in IoT environments. Each method's variability is visualized using shaded regions, representing the standard deviation across multiple runs. Larger uncertainty regions indicate greater sensitivity in counterfactual selection.}\label{fig:Comparison}
\end{figure*}

Figure~\ref{fig:HyperparametersSensitivity} provides an in-depth analysis of the sensitivity of the Free-Energy-based Counterfactual algorithm to variations in the regularization strength, $\lambda$, and the weight of the decision term $\mu$ for different values of the inverse temperature parameter $\beta$. As it can be observed, when $\beta$ decreases, the landscape of stable counterfactual solutions expands, suggesting a broader range of feasible $(\lambda, \mu)$
values that lead to stable explanations. This behaviour aligns with  theoretical properties of the free-energy formulation, where for $\beta\to 0$ values, the entropy term in Eq.~\eqref{eqn:FreeEnergy} dominates, encouraging greater exploration of the solution space. The effect can be also understtod through the Hessian analysis of the the Free-Enrgey fuction. Expanding $\mathcal{F}_{\beta}(\Delta x)$ around an optimal counterfactual $\Delta x^*$ using a second-order Taylor approximation, we get 
\begin{eqnarray}
\mathcal{F}_{\beta}(\Delta \bm{x}) &\approx& \mathcal{F}_{\beta}(\Delta \bm{x}^*) + \nabla_{\bm{x}} \mathcal{F}_{\beta}(\Delta \bm{x}^*) (\Delta \bm{x} - \Delta \bm{x}^*) \nonumber\\ &+& \frac{1}{2} (\Delta \bm{x} - \Delta \bm{x}^*)^{\top} H_{\mathcal{F}_{\beta}} (\Delta \bm{x} - \Delta \bm{x}^*),
\end{eqnarray}
where the Hessian is given by $H_{\mathcal{F}_{\beta}} = H_{\mathcal{E}} - \frac{1}{\beta} H_{\mathcal{S}_{\beta}}$. 

Since $H_{\mathcal{S}_{\beta}}$ represents the entropy contribution, for small $\beta$, $\frac{1}{\beta} H_{\mathcal{S}_{\beta}}$ dominates, reducing the eigenvalues of the Hessian and flattening the optimization landscape. This flattening allows for a larger number of stable counterfactual solutions, leading to a greater variety of valid hyperparameter pairs $(\lambda, \mu)$ that satisfy the counterfactual conditions. Empirically, this effect is reflected in the heatmaps, where lower $\beta$ values result in more regions with high stability scores, indicating that more hyperparameter combinations lead to robust counterfactuals.

This theoretical insight explains why, as $\beta \to 0$, the colormap exhibits a richer variety of valid solutions compared to larger $\beta$ values. When entropy dominates, the optimization process no longer seeks a single minimal perturbation but instead identifies a broader range of plausible perturbations that align with the model’s decision boundary. This increased flexibility enhances robustness but also introduces trade-offs: while the algorithm is less sensitive to precise hyperparameter tuning, it may generate counterfactuals with slightly larger perturbations. Understanding this trade-off is crucial in selecting an appropriate $\beta$ that balances diversity and interpretability in counterfactual generation.

Conversely, for higher values of $\beta$, the stability region shrinks, indicating that the algorithm becomes more selective in finding counterfactuals that satisfy the constraints. This is expected since higher $\beta$ values correspond to a more deterministic optimization process, where the algorithm strongly favors solutions that minimize the free-energy objective rather than exploring alternative plausible explanations. 

The results also reveal an interaction effect between $\lambda$ and $\mu$. In all cases, moderate values of $\lambda$ and $\mu$ yield the highest stability scores, confirming that balancing perturbation regularization and decision boundary alignment is critical for robust counterfactual generation. If $\lambda$ is too small, counterfactuals may become unrealistic due to excessive changes, whereas large $\lambda$ values may prevent meaningful perturbations, leading to ineffective explanations. Similarly, if $\mu$ is too small, the counterfactuals might not sufficiently move across the decision boundary, while excessively large $\mu$ values might cause unnecessary distortions.

\section{Comparison with Other Methods}\label{Sec:CompOther}
To evaluate the effectiveness of counterfactual explanation techniques in IoT cybersecurity, we compare three methods: Free-Energy Counterfactuals, LIME, and Stochastic Gradient-Based Counterfactuals~\cite{Wang:2024}. The objective is to identify minimal but actionable feature perturbations that can flip the model’s classification of an anomalous IoT device state that produces a security alert, to a normal state. Table~\ref{tab:counterfactual_comparison} presents the original input values corresponding to an anomalous security event, serving as the reference for generating counterfactual explanations, along with a comparative evaluation of different counterfactual explanation methods. 

The stochastic gradient-based counterfactual method formulates counterfactual generation as an optimization problem similar to Eq~\ref{eq:cf_baseline}, where the goal is to find a perturbation $\Delta\bm{x}$ that minimally alters the model’s prediction while ensuring feasibility. To solve this optimization problem, it iteratively updates the perturbation based on 
\begin{equation}
\Delta\bm{x}^{(t+1)} = \Delta\bm{x}^{(t)} - \alpha\nabla_{\bm{x}}\mathcal{L}\Big(f(\bm{x}+\Delta\bm{x}),y_{\text{target}}\Big),
\end{equation}
where $\alpha$ is the learning rate, $\mathcal{L}(\cdot)$ is the loss function ensuring the new prediction flips to $y_{\text{target}}$ and $\nabla_{\bm{x}}\mathcal{L}(\cdot)$ is the gradient of the loss function with respect to the input features.

\begin{figure*}[t]
\includegraphics[width=\textwidth]{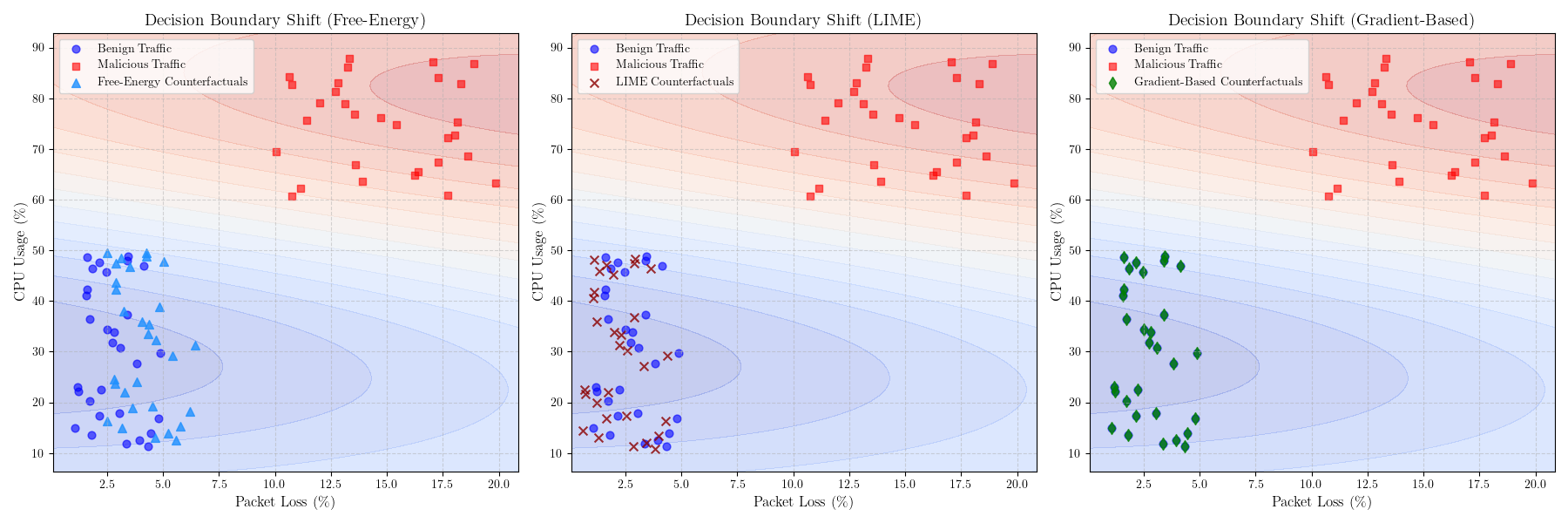}
\caption{Visualization of the counterfactual decision boundary shift for the three methods used for counterfactual explanations in IoT security applications. The contours in each subplot represent the classifier's decision function, where red regions indicate predictions of malicious traffic and blue regions correspond to benign traffic.}\label{fig:DecisionBoundaShift}
\end{figure*}

On the other hand, LIME constructs an interpretable local model $g(\bm{x})$ by solving the following optimization problem
\begin{equation}
\arg\min_{g\in\mathrm{G}}\sum\limits_{i=1}^{N}\pi_{\bm{x}}\left(\bm{x}_{i}\right)\cdot\Big(f(\bm{x}_{i})-g(\bm{x}_{i})\Big)^{2}+\Omega(g)
\end{equation}
where $\mathrm{G}$ is the class of interpretable models, $\pi_{\bm{x}}\left(\bm{x}_{i}\right)$ is a local weighting function, ensuring that perturbed samples closer to $\bm{x}$ have higher influence, $\Omega(g)$ is a complexity term that encourages simpler models and $f\left(\bm{x}_{i}\right)$ is the original Machine Learning model’s output. From the formulation above, it can be easily seen that LIME approximates the model's decision surface, in contrast with the free energy-based counterfactuals that directly optimize perturbations to generate actionable explanations. Table~\ref{tab:method_comparison} lists the optimization objectives and computational properties of these three methods and offers a structured comparison of their foundational principles and practical implications derived from the anlaysis below.

The results of the effectiveness of these three methods are presented in Figure~\ref{fig:Comparison}, which shows the magnitude of perturbation required for each IoT security feature across multiple runs. The $y$-axis represents the magnitude of perturbation required for each feature to generate a counterfactual, while the $x$-axis lists various IoT security metrics used in our scenario. Each method’s variability is illustrated using a shaded region, representing the standard deviation over multiple runs, providing insights into the stability and sensitivity of counterfactual explanations.

The Free-Energy method exhibits the highest variability across different IoT security features. This is expected, as the method introduces a probabilistic energy minimization approach, where perturbations are guided by both the system’s constraints and a simulated annealing process. The larger shaded area suggests that for certain security-critical features (e.g., firmware age, open ports, and packet loss), the counterfactuals require significantly different perturbations in different runs. This variability can be attributed to multiple low-energy solutions, meaning that the algorithm can find different but equally valid counterfactuals that flip the model’s decision.

On the other hand, the LIME method remains almost flat across all features. This indicates that LIME produces highly stable and deterministic perturbations, meaning that for each feature, the generated counterfactual perturbation is nearly the same across multiple runs. This stability stems from LIME’s inherent mechanism since it builds a local interpretable model around the original input using small, linear feature perturbations. Since LIME does not incorporate an optimization-based exploration process, it is constrained to locally linear approximations, resulting in minimal variability across runs.

The Gradient-Based method shows a behavior that lies between the Free-Energy and LIME approaches. The perturbations remain relatively stable but exhibit some degree of variability, particularly in features like port scans and CPU usage. This can be explained by the sensitivity of gradient-based updates to the local landscape of the classifier’s decision boundary. Unlike LIME, which is purely linear, stochastic gradient-based counterfactuals follow the steepest direction of model sensitivity, which can lead to different perturbations across multiple runs due to initialization effects or local minima.

Figure~\ref{fig:DecisionBoundaShift} illustrates how counterfactual explanations generated by the Free-Energy, LIME and Gradient-Based methods affect the decision boundary of a Support Vector Machine (SVM) classifier~\cite{Alex:2023} trained on an IoT security dataset. The $x$-axis represents Packet Loss ($\%$), and the $y$-axis represents CPU Usage ($\%$), two critical features in the IoT network security. The background contours depict the classifier's decision function, with red regions indicating areas classified as malicious and blue regions as benign.

As we observe, the Free-Energy-based counterfactuals exhibit structured and smooth movement towards the decision boundary, with counterfactual samples consistently placed in regions that maximize separability. This aligns with our claims that the Free-Energy framework ensures stability and robustness against local minima, making it a reliable tool for counterfactual generation.

In contrast, LIME generates less structured counterfactuals with greater variance in placement, often scattered around the boundary. This behavior stems from LIME's reliance on local approximations, which may not always generalize well across different regions of the decision space. While LIME provides interpretable counterfactuals, its variability indicates that it may be less robust in high-stakes security applications where stable explanations are required.

The Gradient-based counterfactuals demonstrate a more constrained movement (as also seen in Fig.~\ref{fig:Comparison}), typically staying closer to the original benign samples. While this method is computationally efficient, its limited exploratory capability restricts its ability to generate diverse counterfactual explanations, potentially missing optimal perturbations required to shift samples across the decision boundary. This behaviour, confirms our claim that gradient-based approaches suffer from poor global exploration, which may lead to suboptimal counterfactual suggestions.

\begin{table*}[t]
    \centering
    \renewcommand{\arraystretch}{1.2}
    \caption{Original Input and Counterfactual Explanations for Different Methods}
    \label{tab:counterfactual_comparison}
    \begin{tabular}{|l|c|p{5.8cm}|c|c|c|}
        \hline
        \textbf{Feature} & \textbf{Original Value} & \textbf{Description} & \textbf{Free-Energy} & \textbf{LIME} & \textbf{Gradient-Based} \\ \hline
        \textbf{Packet Loss (\%)} & 12.3 & High packet loss indicating potential DoS attack. & 12.3 $\pm$ 0.02 & 12.3 $\pm$ 0.005 & 12.3 $\pm$ 0.01 \\ \hline
        \textbf{Latency (ms)} & 180 & Increased delay in communication. & 180 $\pm$ 0.015 & 180 $\pm$ 0.003 & 180 $\pm$ 0.01 \\ \hline
        \textbf{Throughput (Mbps)} & 0.5 & Reduced data transmission rate. & 0.5 $\pm$ 0.02 & 0.5 $\pm$ 0.002 & 0.5 $\pm$ 0.00 \\ \hline
        \textbf{Failed Auth Attempts} & 20 & Multiple failed login attempts suggesting brute-force attack. & 20 $\pm$ 0.018 & 20 $\pm$ 0.001 & 20 $\pm$ 0.008 \\ \hline
        \textbf{Port Scans} & 15 & High number of scans signaling reconnaissance activity. & 15 $\pm$ 0.025 & 15 $\pm$ 0.004 & 15 $\pm$ 0.02 \\ \hline
        \textbf{CPU Usage (\%)} & 85 & Elevated usage due to suspicious processes. & 85 $\pm$ 0.01 & 85 $\pm$ 0.002 & 85 $\pm$ 0.00 \\ \hline
        \textbf{Memory Usage (\%)} & 92 & High memory consumption possibly linked to malware. & 92 $\pm$ 0.03 & 92 $\pm$ 0.004 & 92 $\pm$ 0.005 \\ \hline
        \textbf{Battery Drain Rate} & 1.2 & Fast battery depletion, potential malware infection. & 1.2 $\pm$ 0.008 & 1.2 $\pm$ 0.002 & 1.2 $\pm$ 0.007 \\ \hline
        \textbf{Device Temperature (°C)} & 78 & Unusual heat signature due to excessive background processes. & 78 $\pm$ 0.012 & 78 $\pm$ 0.001 & 78 $\pm$ 0.009 \\ \hline
        \textbf{Open Ports} & 8 & Multiple open ports increasing attack surface. & 8 $\pm$ 0.025 & 8 $\pm$ 0.005 & 8 $\pm$ 0.012 \\ \hline
        \textbf{Firmware Age (days)} & 350 & Outdated firmware, potential vulnerability. & 350 $\pm$ 0.07 & 350 $\pm$ 0.002 & 350 $\pm$ 0.01 \\ \hline
        \textbf{Encryption Status} & 0 & No encryption detected, data at risk. & 0 $\pm$ 0.022 & 0 $\pm$ 0.000 & 0 $\pm$ 0.01 \\ \hline
    \end{tabular}
\end{table*}

\section{Computational Complexity Analysis}\label{SecCompComplex}
The computational efficiency of the Free-Energy-Based Counterfactual Algorithm is extremelly critical for its applicability in high-dimensional decision-making systems. We analyze the complexity of key computational steps, derive theoretical bounds and compare our method with the other counterfactual explanation approaches, i.e., LIME and the stochastic gradient-based methods. Its complexity is primarily determined by the computation of the free energy function, the evaluation of the gradient, the computation of the Hessian, the simulated annealing acceptance criterion along with the iteration updates, and last, the convergence properties.

If we denote $d$, the feature dimension, the computation of the free energy function, $\mathcal{F}_{\beta}(\bm{x})$, based on Eq.~\eqref{eqn:FreeEnergy} requires $\mathcal{O}(d)$ operations for the computation of the $ \|\bm{x} - \bm{x}_0\|_{\ell_{2}}$ norm and an additional $\mathcal{O}(d)$ for the evaluation of the regularization function $R(\bm{x})$. The entropy term, $\mathcal{S}_{\beta}(\bm{x})$, can be computed either analytically or numerically using Monte Carlo methods~\cite{Binder:2010}, depending on how $p(\Delta\bm{x})$ is estimated. In the first case, if we assume a Gaussian distribution for $p(\Delta\bm{x})$, i.e., $p(\Delta\bm{x})\sim\mathcal{N}(0,\Sigma)$, the entropy has a closed-form solution
\begin{equation}
\mathcal{S}_{\beta}(\bm{x}) \approx\frac{1}{2}\log\big[\det(\Sigma)\big].
\end{equation}
The complexity of computing the determinant depends on the structure of the covariance matrix $\Sigma$. If $\Sigma$ is diagonal, then $\det(\Sigma)$  is computed in $\mathcal{O}(d)$, leading to an overall entropy computation complexity of $\mathcal{O}(d)$. However, if a full covariance matrix is used, then computing $\det(\Sigma)$ requires matrix inversion, which scales as $\mathcal{O}(d^{3})$ due to the required Cholesky decomposition or eigenvalue decomposition~\cite{Golub:1996}.

In the case where the entropy is computed via sampling (Monte Carlo estimation), then
\begin{equation}
\mathcal{S}_{\beta}(\bm{x}) \approx \frac{1}{K}\sum\limits_{1}^{K}\ln\big[ p(\Delta\bm{x}^{(i)}) \big].
\end{equation}
For $K$ samples, $\mathcal{O}(Kd)$ operations are required, which is $\mathcal{O}(d)$ or a small, constant $K$, but scales poorly if high precision is needed. Under the typical assupmption for high-dimensional problems that $K=\mathcal{O}(Kd)$, the total complexity of Monte Carlo estimation becomes $\mathcal{O}(d^{2})$. Thus, the overall complexity of computing $\mathcal{F}_{\beta}(\bm{x})$ depends on the entropy estimation method. In the best case, when a diagonal covariance Gaussian assumption or efficient entropy estimation is used, the complexity is $\mathcal{O}(d)$. However, if Monte Carlo entropy estimation with a full covariance model is employed, the complexity increases to $\mathcal{O}(d^{2})$ or even $\mathcal{O}(d^{3})$ with full covariance inversion.

The computation of the gradient $\nabla_{\bm{x}} \mathcal{F}_{\beta}(\bm{x})$ strongly contributes to the iterative refinement process of our Free-Energy-based Counterfactual algorithm, since it determines the direction in which perturbations should be applied to achieve minimal modifications while ensuring the counterfactual validity. Its computational complexity can be derived analyzing the gradients of the components of $\mathcal{F}_{\beta}(\bm{x})$ based on Eq.~\eqref{eqn:FreeEnergy}. Starting with the gradient of $\mathcal{E}(\bm{x})$
\begin{equation}\label{eqn:GradEner}
    \nabla_{\bm{x}} \mathcal{E}(\bm{x}) = \frac{\bm{x} - \bm{x}_0}{\|\bm{x} - \bm{x}_0\|_2} + \lambda \nabla_{\bm{x}} R(\bm{x}) + \mu \nabla_{\bm{x}} | f(\bm{x}) - c |.
\end{equation}
To further simplify the gradient computation, we apply a first-order Taylor expansion to approximate the model’s decision function $f(\bm{x})$ in a small region around $\bm{x}$ as
\begin{equation}
f(\bm{x} + \Delta \bm{x}) \approx f(\bm{x}) + \nabla f(\bm{x})^{\top} \Delta \bm{x}.
\end{equation}
Next, we take the gradient of the decision boundary alignment term, $| f(\bm{x}) - c |$ and use the sign function, since the absolute value function is non-differentiable at zero,
\begin{equation}
    \nabla_{\bm{x}} | f(\bm{x}) - c | = \text{sign}(f(\bm{x}) - c) \nabla f(\bm{x}).
\end{equation}
The obtained result reveals that if $f(\bm{x}) > c$, the function is increasing, and the perturbation should move against the gradient of $f(\bm{x})$. In cotrnast, if $f(\bm{x}) < c$, the function is decreasing, and the perturbation should move in the direction of the gradient of $f(\bm{x})$. Thus, substituting this result into the gradient expression in Eq.~\eqref{eqn:GradEner}, we obtain:
\begin{equation}
    \nabla_{\bm{x}} \mathcal{E}(\bm{x}) = \frac{\bm{x} - \bm{x}_0}{\|\bm{x} - \bm{x}_0\|_2} + \lambda \nabla_{\bm{x}} R(\bm{x}) + \mu \, \text{sign}(f(\bm{x}) - c) \nabla f(\bm{x}).
\end{equation}
The computational cost of evaluating this gradient can be splitted into three main parts. For the computation of $\frac{\bm{x} - \bm{x}_0}{\|\bm{x} - \bm{x}_0\|_2}$, $\mathcal{O}(d)$ operations are required. The same holds for the evaluation of $\nabla_{\bm{x}} R(\bm{x})$ and the computation of the  classifier gradient $\nabla f(\bm{x})$ (if it is differntaible).
Thus, the total complexity of $\nabla_{\bm{x}} \mathcal{E}(\bm{x})$ remains $\mathcal{O}(d)$ which ensures that gradient updates remain computationally efficient, making the algorithm scalable for high-dimensional feature spaces.

The computation of the entropy gradient $\nabla_{\bm{x}} \mathcal{S}_{\beta}(\bm{x})$ reflects the balance of exploration and minimization within the free-energy framework. Since entropy is defined as an integral over the perturbation space, its exact computation is often intractable, requiring numerical approximations~\cite{Noughabi:2015}. For this purpose, we use a Monte Carlo sampling method, where $K$ perturbation samples $\{ \Delta \bm{x}_i \}_{i=1}^{K}$ are drawn from the probability distribution $p(\Delta \bm{x})$. The gradient estimate is given by:
\begin{eqnarray}
\nabla_{\bm{x}} \mathcal{S}(\bm{x}) &\approx& -\sum_{i=1}^{K}\Big[\nabla_{\bm{x}} p(\Delta \bm{x}_i)\ln\left[p(\Delta \bm{x}_i)\right]\nonumber\\
&+& \nabla_{\bm{x}} p(\Delta \bm{x}_i) \Big].
\end{eqnarray}
The complexity of the entropy gradient computation depends on multiple factors, including the sampling process, the probability density function,  the gradient computation and the summation over the samples.

The first factor influencing complexity is the sampling process. Generating $K$ perturbation samples from the distribution $p(\Delta \bm{x})$ typically requires $\mathcal{O}(K)$ operations. In the case of simple distributions such as a Gaussian perturbation model, each sample can be drawn in constant time. However, for more complex non-parametric distributions, such as those estimated via kernel density estimation or neural-based probabilistic models, sampling may involve Markov Chain Monte Carlo methods, increasing the cost to $\mathcal{O}(Kd)$.

Once samples are generated, each perturbation $\Delta \bm{x}_{i}$ requires evaluating its probability density function, $p(\Delta \bm{x}_{i})$. The complexity of this step depends on the underlying form of $ p(\Delta \bm{x}) $. For a Gaussian distribution, each evaluation requires $\mathcal{O}(d)$ operations. In the case of kernel density estimation, the computational cost increases to $\mathcal{O}(Kd)$, since it involves evaluating each sample against the entire dataset. If a neural-based density model is used, the complexity can be as high as $\mathcal{O}(d^{2})$, depending on the architecture of the density estimator. Given these variations, the best-case complexity for PDF evaluation is $\mathcal{O}(Kd)$.

In addition to the $p(\Delta \bm{x})$ evaluation, computing the gradient $\nabla_{\bm{x}} p(\Delta \bm{x}_i)$ for each perturbation sample contributes further complexity. If the perturbation distribution follows a Gaussian model with mean $\mathbb{E}[\bm{x}]$ and covariance matrix $\Sigma$, then the gradient can be computed as $ \nabla_{\bm{x}} p(\Delta \bm{x}) = p(\Delta \bm{x}) \Sigma^{-1} (\Delta \bm{x} - \bm{x}) $. The key computational bottleneck here is the inversion of $ \Sigma $, which, in the general case (as also described above), has a complexity of $\mathcal{O}(d^{3})$. If the inverse is precomputed and reused across iterations, the complexity is reduced to $\mathcal{O}(d^{2})$, yielding an overall cost of $\mathcal{O}(Kd^{2})$ for computing all gradients.

Following these computations, the entropy gradient estimation requires summing over all $K$ Monte Carlo samples. Since this operation involves a simple element-wise addition, it contributes an additional $\mathcal{O}(Kd)$ operations, which is generally negligible compared to the cost of probability evaluations and gradient computations.

Summing all these components, the total complexity of computing the entropy gradient is $\mathcal{O}(Kd^{2})$. However, in cases where the covariance matrix $\Sigma$ is precomputed, the computational cost reduces to $\mathcal{O}(Kd)$. The choice of $K$ significantly affects both computational cost and accuracy. Increasing $K$ leads to a more accurate entropy gradient estimate by reducing variance but also increases computational cost linearly. For high-dimensional settings, using a precomputed covariance matrix can significantly reduce complexity while maintaining efficiency. In practical implementations, $K$ is typically set to a moderate value to strike a balance between computational feasibility and accuracy in entropy estimation.

The iterative nature of the optimization process introduces an additional dependency on the number of iterations $T$ required for convergence. Under the assumption that the number of iterations scales as $\mathcal{O}(T)$ the total complexity of the Free-Energy-Based Counterfactual Algorithm is $\mathcal{O}(Td)$ when entropy estimation is efficiently handled and $\mathcal{O}(TKd^{2})$ when Monte Carlo-based entropy estimation with full covariance modeling is employed. In practical scenarios, where $K$ is chosen to be moderate and covariance structures are precomputed, the expected complexity remains closer to $\mathcal{O}(TKd)$, making our method scalable for high-dimensional counterfactual search problems while ensuring efficient convergence.

To assess the computational efficiency of the proposed Free-Energy-Based Counterfactual Algorithm, we compare its complexity with two widely used counterfactual explanation methods, i.e., LIME and Gradient-Based Counterfactual Search~\cite{Han:2023}. The complexity analysis is performed in terms of the feature dimension $d$, the number of samples $K$, and the number of iterations $T$, which directly impact the computational feasibility of each method.

\begin{figure}[t]
\includegraphics[width=0.5\textwidth]{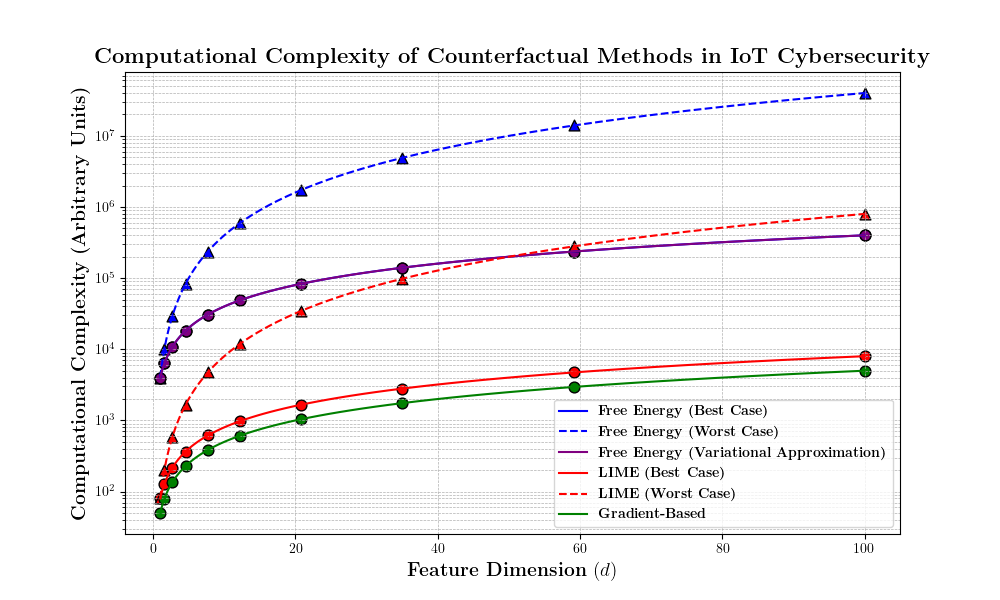}
\caption{Computational complexity comparison of the counterfactual explanation methods including their best and worst cases.}\label{fig:ComplexityComparison}
\end{figure}

LIME approximates the local decision boundary of a model by training a surrogate model on perturbed data points. This requires generating $K$ perturbed samples, querying the black-box model for predictions and solving a weighted regression problem to approximate the local behavior of the decision function. The most computationally intensive steps in LIME include the generation of perturbations, which scales as $\mathcal{O}(K d)$, and the evaluation of the model predictions for each sample, which requires $\mathcal{O}(K)$ operations. The most compurtationally intensive component is the fitting of a surrogate model, where solving a weighted regression problem results in a complexity of $\mathcal{O}(K d^{2})$. Since LIME typically requires a large number of samples (e.g., $K > 5000$) to obtain stable explanations, its computational cost is considerably high for high-dimensional feature spaces. Unlike iterative methods, LIME has a fixed runtime for a given $K$, but this comes at the expense of requiring a large number of perturbed samples to accurately approximate the local decision boundary.

Gradient-Based Counterfactual Search directly optimizes an objective function, minimizing the distance to the original instance while ensuring a prediction change. This is typically achieved using gradient descent, which consists of computing $\nabla_{\bm{x}} f(\bm{x})$ in each iteration. The computation of the gradient requires $\mathcal{O}(d)$ operations per update step, followed by a first-order optimization update with the same complexity. The number of iterations $T$ required for convergence determines the total runtime of the algorithm. Thus, the overall complexity of gradient-based counterfactual search is $\mathcal{O}(T d)$, making it computationally efficient. However, such methods are highly sensitive to local minima, requiring additional constraints or optimization techniques to improve robustness.

As observed in Fig.~\ref{fig:ComplexityComparison}, LIME exhibits the highest computational cost due to the extensive number of perturbations required and the subsequent weighted regression fitting, leading to a worst-case complexity of $\mathcal{O}(K d^{2})$. In contrast, gradient-based counterfactual methods are the most computationally efficient, with a complexity of $\mathcal{O}(T d)$, but they often struggle with poor global exploration, making them susceptible to local minima and unstable counterfactual generation. The Free-Energy-Based Counterfactual Algorithm achieves a balance between these two approaches by incorporating simulated annealing and entropy preservation, enhancing robustness against local optima. While its computational demand is higher than gradient-based methods, efficient entropy estimation and a moderate choice of $K$ keep its expected complexity at $\mathcal{O}(T K d)$, allowing it to scale effectively in high-dimensional counterfactual search. Consequently, Free-Energy-based counterfactuals offer more refined and globally stable explanations, albeit at a significantly higher computational cost compared to purely gradient-based approaches.

\begin{table*}[t]
    \centering
    \renewcommand{\arraystretch}{1.2}
    \caption{Comparison of Free-Energy-Based Counterfactuals, LIME, and Stochastic Gradient-Based Counterfactuals in IoT Security.}
    \label{tab:method_comparison}
    \begin{tabular}{|p{2.2cm}|p{4.7cm}|p{4.7cm}|p{4.7cm}|}
        \hline
        \textbf{Aspect} & \textbf{Free-Energy Counterfactuals} & \textbf{LIME} & \textbf{Stochastic Gradient Counterfactuals} \\
        \hline
        \textbf{Optimization Goal} & 
        Find minimal perturbations $\Delta \bm{x}$ to alter model prediction while balancing perturbation cost and entropy for robustness. & 
        Fit a local interpretable model $g(\bm{x})$ around an input instance for feature attribution & 
        Solve $\min\limits_{\Delta \bm{x}} \|\Delta \bm{x}\|$ subject to flipping the model decision using gradient updates. \\
        \hline
        \textbf{Mathematical Formulation} & 
        $\mathcal{F}_{\beta}(\Delta \bm{x}) = \mathcal{E}(\Delta \bm{x}) - \frac{1}{\beta} \mathcal{S}(\Delta \bm{x})$ with simulated annealing for optimization. & 
        $\sum_{i} \pi_{\bm{x}}(\bm{x}_i) \cdot (f(\bm{x}_i) - g(\bm{x}_i))^2 + \Omega(g)$, where $\pi_{\bm{x}}$ is a local weighting function. & 
        $\Delta \bm{x}^{(t+1)} = \Delta \bm{x}^{(t)} - \alpha \nabla_{\bm{x}} \mathcal{L}(f(\bm{x} + \Delta \bm{x}), y_{\text{target}})$. \\
        \hline
        \textbf{Energy Term $\mathcal{E}$} & Encodes perturbation cost, security constraints and decision boundary alignment. & Measures model approximation error through local surrogate models. & Direct gradient update towards crossing the decision boundary. \\
        \hline
        \textbf{Entropy Term $\mathcal{S}$} & Ensures counterfactual diversity and robustness to adversarial perturbations. & Not explicitly modeled; relies on smoothness assumptions in local approximation. & Implicit randomness due to stochastic gradient updates but lacks entropy modeling. \\
        \hline
        \textbf{Constraint Handling} & Hard feasibility constraints for the features; penalizes changes to critical variables. & Regularization $\Omega(g)$ for local model simplicity but does not impose explicit feature constraints. & Soft constraints using gradient-based updates; struggles with strict feature constraints in high-stakes security applications. \\
        \hline
        \textbf{Computational Complexity} & Moderate-to-high: $\mathcal{O}(T K d)$ in the best case with efficient entropy estimation; can reach $\mathcal{O}(T K d^2)$ in worst-case Monte Carlo sampling. & Low: $\mathcal{O}(K d^2)$ due to fitting a local model with weighted regression. & Low-to-moderate: $\mathcal{O}(T d)$, but sensitive to initialization and requires many iterations for stability. \\
        \hline
        \textbf{Interpretability} & Produces minimal, plausible counterfactuals, optimized for various settings. & Provides feature importance explanations, but does not generate direct counterfactual samples. & Generates perturbations efficiently, but lacks interpretability beyond sensitivity analysis. \\
        \hline
        \textbf{Robustness} & High: Incorporates entropy regularization, avoiding adversarial drift and ensuring stable counterfactual selection. & Low: LIME’s local linear approximations are highly sensitive to adversarial manipulations. & Medium to High: Gradient-based methods can be adversarially manipulated but remain stable for well-conditioned optimization problems. \\
        \hline
        \textbf{Scalability to High-Dimensional Data} & High: Free-energy minimization framework scales well with entropy-based search strategies, mitigating curse of dimensionality. & Low: Becomes infeasible for high-dimensional data due to reliance on local surrogate models. & High: Gradient-based search is efficient, but optimization can become unstable in very high-dimensional spaces. \\
        \hline
    \end{tabular}
\end{table*}

There are several techniques that can be employed to lower the overall complexity of the Free-Energy algorithm, focusing on the entropy term which is the primary contributor to the computational overhead. For example, instead of computing directly the entropy, we can employ variational methods in order to approximate it. In this case we introduce a tractable surrogate function $q(\Delta \bm{x})$ that is computationally efficient and through the Kullback-Leibler (KL) divergence~\cite{Kullback:1951}, we approximate entropy in Eq.~\eqref{eqn:EntropyTerm} as:
\begin{equation}\label{eqn:EntropyTermGauss}
\mathcal{S}_{\beta} = -\int p(\Delta \bm{x}) \ln \Big[q(\Delta \bm{x})\Big]\mathrm{d}\Delta \bm{x} + \mathcal{D}_{\mathrm{KL}}(p||q), 
\end{equation}
where the KL divergence  
\begin{equation}
\mathcal{D}_{\mathrm{KL}}(p||q) = \int p(\Delta \bm{x}) \ln \Bigg[\frac{p(\Delta \bm{x})}{q(\Delta \bm{x})}\Bigg]\mathrm{d}\Delta \bm{x},
\end{equation}
quantifies the discrepancy between the true distribution $p(\Delta \bm{x})$ and the variational approximation $q(\Delta \bm{x})$. If weuse a Gaussian approximation, i.e., $q(\Delta \bm{x})\sim\mathcal{N}(\mathbb{E}[\bm{x}], \Sigma)$, the entropy has a closed-form expression
\begin{equation}
\mathcal{S}_{\beta} \approx \frac{1}{2}\ln\big[\det (\Sigma)\big] + \frac{d}{2}\ln [2\pi e].
\end{equation}
Using Jensen’s inequality, we can derive an upper bound on the entropy,
\begin{equation}
\mathcal{S}_{\beta} \geq -\mathbb{E}_{q}\ln\big[q(\Delta\bm{x})\big] - \mathcal{D}_{\mathrm{KL}}(p||q),
\end{equation}
ensuring that even if $q(\Delta \bm{x})$ is a rough approximation, the overall estimation remains valid. Moreover, if the KL divergence is small, then $q(\Delta \bm{x})$ serves as a near-optimal approximation of $p(\Delta \bm{x})$.

The adoption of the Gaussian variational approximation  in Free-Energy counterfactuals can reduce the entropy complexity to $\mathcal{O}(d)$ (assuming feature independence), providing a computationally feasible alternative to the computation of the entropy term. The replacement of the computational intensive Monte Carlo sampling and matrix inversion with closed-form Gaussian approximation can achieve an order-of-magnitude reduction. Despit ethe introduction of slight approximation errors, these are well-controlled through variational bounds, ensuring robustness while preserving interpretability. 

The variational approach can lead to a complexity similar to the best case scenario of the Free-Energy algorithm where the covariance matrix $\Sigma$ is assumed to be diagonal, allowing entropy computations to be performed in $\mathcal{O}(d)$.  This allows entropy computations to scale linearly with feature dimensionality, making it suitable for high-dimensional applications where real-time counterfactual generation is a key factor. Additionally, variational approximations retain the ability to model uncertainty and diversity in counterfactuals while avoiding the computational bottlenecks associated with more complex entropy estimations.   

\section{Robustness of the Free-Energy-based Framework}\label{Sec:Robust}
The robustness of the Free-Energy-Based Counterfactual Framework hinges on the stability of the free energy function $\mathcal{F}_{\beta}\big(\bm{x}\big)$ when subjected to small perturbations. To ensure counterfactual stability, we examine the Lipschitz continuity, the gradient behavior and the Hessian spectral properties of $\mathcal{F}_{\beta}\big(\bm{x}\big)$.

Given a small perturbation $\bm{\eta}$, the free energy function defined in Eq.~\eqref{eqn:FreeEnergy} can be approximated using a first-order Taylor expansion:
\begin{equation}\label{eqn:TayExpann}
\mathcal{F}_{\beta}(\bm{x} + \bm{\eta}) \approx \mathcal{F}_{\beta}(\bm{x}) + \nabla \mathcal{F}_{\beta}(\bm{x})^{\top} \bm{\eta}.
\end{equation}
This approximation suggests that if the gradient norm $\|\nabla \mathcal{F}_{\beta}(\bm{x})\|$ is large, even small perturbations can lead to significant changes in the free energy landscape, potentially causing counterfactual instability. To prevent excessive sensitivity, we impose a Lipschitz continuity condition~\cite{Eriksson:2004}
\begin{equation}
\|\mathcal{F}_{\beta}(\bm{x}_1) - \mathcal{F}_{\beta}(\bm{x}_2)\| \leq L \|\bm{x}_1 - \bm{x}_2\|,
\end{equation}
where $L$ is the Lipschitz constant that bounds how much $\mathcal{F}_{\beta}(\bm{x})$ can change locally. This condition ensures that counterfactuals generated under similar inputs remain stable and do not exhibit abrupt changes in response to small perturbations.

Beyond first-order sensitivity, we also examine the second-order stability of the free energy function, which accounts for local curvature. Using the Hessian matrix,
\begin{equation}
H_{\mathcal{F}_{\beta}}(\bm{x}) = \nabla^2 \mathcal{F}_{\beta}(\bm{x}),
\end{equation}
we expand $\mathcal{F}_{\beta}(\bm{x})$ using a second-order Taylor series as
\begin{equation}
\mathcal{F}_{\beta}(\bm{x} + \bm{\eta}) \approx \mathcal{F}_{\beta}(\bm{x}) + \nabla \mathcal{F}_{\beta}(\bm{x})^{\top} \bm{\eta} + \frac{1}{2} \bm{\eta}^{\top}\big[H_{\mathcal{F}_{\beta}}(\bm{x})\big] \bm{\eta},
\end{equation}
where the spectral properties of the Hessian dictate how sharply the function changes. The largest eigenvalue $\lambda_{\max} (H_{\mathcal{F}_{\beta}}(\bm{x}))$ determines the curvature of $\mathcal{F}_{\beta}(\bm{x})$. For stability, we enforce the constraint:
\begin{equation}
\lambda_{\max} \Big(H_{\mathcal{F}_{\beta}}(\bm{x})\Big) \leq C,
\end{equation}
where $C$ is a small constant that ensures that the counterfactual optimization landscape remains smooth. If the curvature is excessively high, small changes in $\bm{x}$ can lead to disproportionately large deviations in $\mathcal{F}_{\beta}(\bm{x})$, resulting in counterfactuals that are not robust.

In addition to ensuring smooth local behavior, we investigate the robustness of counterfactuals under adversarial perturbations. An adversary may introduce a worst-case perturbation $\bm{\eta}^*$ that maximizes the change in free energy:
\begin{equation}\label{eqn:worpert}
\bm{\eta}^{*} = \arg\max_{\|\bm{\eta}\| \leq \xi} \mathcal{F}_{\beta}(\bm{x} + \bm{\eta}) - \mathcal{F}_{\beta}(\bm{x}),
\end{equation}
where $\xi$ represents the perturbation budget or the magnitude constraint on the adversarial perturbation $\bm{\eta}$. Substituting the Taylor expansion of Eq.~\eqref{eqn:TayExpann} into Eq.~\ref{eqn:worpert} , the worst-case shift in free energy is given by:
\begin{equation}
\max_{\|\bm{\eta}\| \leq \epsilon} \|\nabla \mathcal{F}_{\beta}(\bm{x})\|_{\ell_{2}} \|\bm{\eta}\|_{\ell_{2}} + \frac{1}{2} \lambda_{\max} \Big(H_{\mathcal{F}_{\beta}}(\bm{x})\Big) \|\bm{\eta}\|_{\ell_{2}}^2.
\end{equation}
As it can be readily seen by this formulation, the robustness under adversarial perturbations is governed by both the gradient magnitude and the Hessian curvature. To prevent adversarial vulnerabilities, we impose a robustness condition
\begin{equation}
\lambda_{\max} (H_{\mathcal{F}_{\beta}}) \leq \frac{2}{\epsilon} \|\nabla \mathcal{F}_{\beta}(\bm{x})\|_{\ell_{2}},
\end{equation}
that ensures that even under worst-case perturbations, the counterfactual remains valid and does not exhibit drastic fluctuations.

All these conditions have been taken into consideration as it can be observed from Algorithm~\ref{alg:cf} and the energy landscape visualization in Fig.~\ref{fig:EnergyLandscape} where our method avoids local minima. The incorporation of gradient sensitivity constraints ensures that counterfactual updates remain stable, while the Hessian-based spectral bound prevents extreme curvature fluctuations in the optimization process. Additionally, the simulated annealing framework allows for controlled exploration, enabling the method to escape poor local optima and converge toward minimal yet plausible counterfactuals. The adversarial robustness check further validates that the generated counterfactuals remain valid under small perturbations, demonstrating the overall stability of the approach.

\section{Conclusion and Future Work}\label{Sec:ConFut}
In this paper, we introduced a novel framework that integrates perturbation theory and statistical mechanics to generate minimal, actionable counterfactual explanations. Unlike traditional methods, which rely on direct optimization or heuristic search, our work reframes counterfactual search as an energy minimization problem over a complex landscape. Using local Taylor expansions, we approximate the decision boundary of a Machine Learning model, enabling more efficient counterfactual generation. The employment of simulated annealing that navigates the energy landscape, ensures that the identified counterfactuals are both realistic and robust. The proposed framework was tested in the context of cybersecurity applications in an IoT environment. The experimental results demonstrate that the method outperforms well-established, state-of-the-art methods such as LIME and the Gradient-based Counterfactuals, offering improved interpretability while capturing the model's sensitivity to input perturbations.

Building upon this energy-based counterfactual generation framework, future research can incorporate domain-specific constraints and fairness-aware counterfactuals that can further enhance its applicability to more sophisticated security decision-making systems. Additionally, integrating adaptive cooling schedules in simulated annealing could further optimize the trade-off between exploration and convergence in high-dimensional spaces. Another important issue that it worths exploring is the extension of our framework to generative AI models, where counterfactual generation could improve model alignment and trustworthiness. Lastly, real-world validation in dynamic cybersecurity environments, particularly in adversarial attack detection and mitigation, would provide deeper insights into the framework’s robustness under evolving threat landscapes.

\begin{thebibliography}{00}
\bibitem{Ronge:2024}
R. Ronge, B. Bauer and B. Rathgeber, ``Approaching Principles of XAI: A Systematization," in IEEE Transactions on Artificial Intelligence, vol. 1, no. 01, pp. 1-13, Aug. 2024, doi: 10.1109/TAI.2024.3515937.

\bibitem{Celar:2023}
L.~Celar and R.M.J.~Byrne, ``How People Reason with Counterfactual and Causal Explanations for Artificial Intelligence Decisions in Familiar and Unfamiliar Domains,'' Springer Nature, Memory \& Cognition, vol. 51, pp. 1481--1496, 2023, doi: 10.3758/s13421-023-01407-5.

\bibitem{Stepin:2021}
I. Stepin, J. M. Alonso, A. Catala and M. Pereira-Fariña, ``A Survey of Contrastive and Counterfactual Explanation Generation Methods for Explainable Artificial Intelligence," in IEEE Access, vol. 9, pp. 11974--12001, 2021, doi: 10.1109/ACCESS.2021.3051315.

\bibitem{Mothilal:2021FeatAttrib}
R.~K.~Mothilal, D.~Mahajan, C.~Tan, and A.~Sharma, ``Towards Unifying Feature Attribution and Counterfactual Explanations: Different Means to the Same End,'' in Proceedings of the 2021 AAAI/ACM Conference on AI, Ethics, and Society (AIES '21), Association for Computing Machinery, New York, NY, USA, pp. 652–663, 2021, doi: 10.1145/3461702.3462597

\bibitem{Sacchetti:2009}
A.~Sacchetti, ``Perturbation Theory, Semiclassical,'' in R.~Meyers, (eds) Encyclopedia of Complexity and Systems Science, 2009, Springer, New York, NY, doi: 10.1007/978-0-387-30440-3\_403.

\bibitem{Huang:1987}
K.~Huang, ``Statistical Mechanics,'' John Wiley \& Sons, ISBN: 978-0-471-81518-1.

\bibitem{Rawal:2022}
A.~Rawal, J.~McCoy, D.~B.~Rawat, B.~M.~Sadler and R.~S.~Amant, ``Recent Advances in Trustworthy Explainable Artificial Intelligence: Status, Challenges, and Perspectives,'' in IEEE Transactions on Artificial Intelligence, vol. 3, no. 6, pp. 852-866, Dec. 2022, doi: 10.1109/TAI.2021.3133846.

\bibitem{Wachter:2017}
S.~Wachter, B.~Mittelstadt, and C.~Russell, ``Counterfactual Explanations without Opening the Black Box: Automated Decisions and the GDPR,'' Harv. JL \& Tech., vol. 31, 841, 2017, doi: 10.2139/ssrn.3063289.

\bibitem{Huang:2023}
C.~Huang, Z.~Zhang, B.~Mao and X.~Yao, ``An Overview of Artificial Intelligence Ethics," in IEEE Transactions on Artificial Intelligence, vol. 4, no. 4, pp. 799--819, Aug. 2023, doi: 10.1109/TAI.2022.3194503.


\bibitem{Verma:2024}
S.~Verma, V.~Boonsanong, M.~Hoang, K.~Hines, J.~Dickerson, and C.~Shah, ``Counterfactual Explanations and Algorithmic Recourses for Machine Learning: A Review,'' in ACM Computing Surveys, vol. 56, no. 12, December 2024, pp. 1--42 pages, doi: 10.1145/3677119.

\bibitem{Guidotti:2024}
R.~Guidotti, ``Counterfactual Explanations and How to Find Them: Literature Review and Benchmarking,'' in Springer Data Mining and Knowledge Discovery, vol. 38, pp. 2770--2824, 2024.

\bibitem{Mothilal:2020}
R.~K.~Mothilal, A.~Sharma, and C.~Tan, ``Explaining Machine Learning Classifiers through Diverse Counterfactual Explanations,'' in Proceedings of the 2020 Conference on Fairness, Accountability, and Transparency (FAT* '20), Association for Computing Machinery, New York, NY, USA, 2020, pp. 607--617, doi: 10.1145/3351095.3372850.

\bibitem{Bayrak:2024}
B.~Bayrak and K.~Bach, ``Evaluation of Instance-Based Explanations: An In-Depth Analysis of Counterfactual Evaluation Metrics, Challenges, and the CEval Toolkit,'' in IEEE Access, vol. 12, pp. 137683--137695, 2024, doi: 10.1109/ACCESS.2024.3410540. 

\bibitem{Ustun:2019}
B.~Ustun, A.~Spangher, and Y.~Liu, ``Actionable Recourse in Linear Classification,'' in Proceedings of the Conference on Fairness, Accountability, and Transparency (FAT* '19), Association for Computing Machinery, New York, NY, USA, 2019, pp. 10--19, doi: 10.1145/3287560.3287566.

\bibitem{Delaunay:2022}
J.~Delaunay, L.~Galárraga, and C.~Largouët, ``When Should We Use Linear Explanations?,'' in Proceedings of the 31st ACM International Conference on Information \& Knowledge Management (CIKM '22), Association for Computing Machinery, New York, NY, USA, 355–364, 2022, doi: 10.1145/3511808.3557489.

\bibitem{Russell:2019}
C.~Russell, ``Efficient Search for Diverse Coherent Explanations,'' in Proceedings of the Conference on Fairness, Accountability, and Transparency (FAT* '19),  Association for Computing Machinery, New York, NY, USA, 2019, pp. 20--28, doi: 10.1145/3287560.3287569.

\bibitem{Karimi:2020}
A.~H.~Karimi, G.~Barthe, B.~Balle, and I.~Valera, ``Model-agnostic Counterfactual Explanations for Consequential Decisions,'' in Proceedings of the 23rd International Conference on Artificial Intelligence and Statistics (AISTATS), vol. 108, 2020, pp. 895--905, doi: 10.48550/arXiv.1905.11190.

\bibitem{Leofante:2023}
J.~Jiang, F.~Leofante, A.~Rago, and F.~Toni, ``Formalising the Robustness of Counterfactual Explanations for Neural Networks,'' Proceedings of the AAAI Conference on Artificial Intelligence, vol. 37(12), pp. 14901--14909, doi: 10.1609/aaai.v37i12.26740

\bibitem{Ferrario:2022}
A.~Ferrario and M.~Loi, ``The Robustness of Counterfactual Explanations Over Time,'' in IEEE Access, vol. 10, pp. 82736-82750, 2022, doi: 10.1109/ACCESS.2022.3196917.

\bibitem{Kato:1966}
T.~Kato, ``Perturbation Theory for Linear Operators,'' Springer-Verlag, Berlin, Heidelberg, New York, 1966.

\bibitem{Ivanovs:2021}
M.~Ivanovs, R.~Kadikis, K.~Ozols, ``Perturbation-based Methods for Explaining Deep Neural Networks: A Survey,'' Pattern Recognition Letters, vol. 150, 2021, pp. 228--234, ISSN 0167-8655, doi: 10.1016/j.patrec.2021.06.030.

\bibitem{Ribeiro:2016}
M.~Ribeiro, S.~Singh, and C.~Guestrin, ``Why Should I Trust You?": Explaining the Predictions of Any Classifier,'' in Proceedings of the 22nd ACM SIGKDD International Conference on Knowledge Discovery and Data Mining (KDD '16), Association for Computing Machinery, New York, NY, USA, 2016, pp. 1135–1144, doi: 10.1145/2939672.2939778.

\bibitem{Yang:2023}
W.~Yang, {\it et al.}, ``Survey on Explainable AI: From Approaches, Limitations and Applications Aspects,'' in Springer Human-Centric Intelligent Systems, vol. 3, 2023, pp. 161--188, doi: 10.1007/s44230-023-00038-y.

\bibitem{Evangelatos:2023}
S.~Evangelatos, and A.~L.~Moustakas, ``Detection of Transmission State of Multiple Wireless Sources: A Statistical Mechanics Approach,'' in MDPI Telecom, vol. 4(3), 2023, pp. 649-677, doi: 10.3390/telecom4030029.

\bibitem{Sourlas:1994}
N.~Sourlas, ``Statistical Mechanics and Error-Correcting Codes,'' In P.~Grassberger, JP.~Nadal, (eds), From Statistical Physics to Statistical Inference and Back, NATO ASI Series, vol. 428, Springer, Dordrecht, doi: 10.1007/978-94-011-1068-6.

\bibitem{Decelle:2023}
A.~Decelle, ``An Introduction to Machine Learning: a Perspective from Statistical Physics,'' in Physica A: Statistical Mechanics and its Applications, vol. 631,
2023, 128154, ISSN 0378-4371, doi: 10.1016/j.physa.2022.128154.

\bibitem{Krzakala:2024}
F.~Krzakala and L.~Zdeborová, ``Statistical Physics Methods in Optimization and Machine Learning.'' Lecture Notes, 2024.

\bibitem{Jin:2017:saddle}
C.~Jin, R.~Ge, P.~Netrapalli, S.~M.~Kakade and M.~I.~Jordan, ``How to Escape Saddle Points Efficiently,'' in Proceedings of the 34th International Conference on Machine Learning, vol. 70, pp. 1724--1732, 2017.

\bibitem{Wang:2021}
C.~Wang, X.-H.~Li, H.~Han, S.~Wang, L.~Wang, C.~C.~Cao, and L.~Chen, ``Counterfactual Explanations in Explainable AI: A Tutorial,'' in Proceedings of the 27th ACM SIGKDD Conference on Knowledge Discovery \& Data Mining (KDD '21), Association for Computing Machinery, New York, NY, USA, 2021, pp. 4080--4081,  doi: 10.1145/3447548.3470797.

\bibitem{Lundberg:2017:SHAP}
S.~M.~Lundberg and S.-I.~Lee, ``A Unified Approach to Iinterpreting Model Predictions,'' in Proceedings of the 31st International Conference on Neural Information Processing Systems (NIPS'17), Curran Associates Inc., Red Hook, NY, USA, 2017, pp. 4768–4777.

\bibitem{Wang:2024}
X.~Wang, Q.~Li, D.~Yu, Q.~Li, and G.~Xu, ``Counterfactual Explanation for Fairness in Recommendation,'' in ACM Transactions on Information Systems, vol. 42, no. 4, Article 106, July 2024, doi: 10.1145/3643670.

\bibitem{Binder:2010}
K.~Binder and D.~W.~Heermann, ``Monte Carlo Methods for the Sampling of Free Energy Landscapes,'' in: Monte Carlo Simulation in Statistical Physics, Graduate Texts in Physics, vol 0, 2010, Springer, Berlin, Heidelberg, doi: 10.1007/978-3-642-03163-2\_6

\bibitem{Alex:2023}
C.~Alex, G.~Creado, W.~Almobaideen, O.~A.~Alghanam and M.~Saadeh, ``A Comprehensive Survey for IoT Security Datasets Taxonomy, Classification and Machine Learning Mechanisms,'' in Computers \& Security, vol. 132, 2023, 103283, ISSN 0167-4048, doi: 10.1016/j.cose.2023.103283.

\bibitem{Golub:1996}
G.~H.~Golub and C.~F.~Van Loan, ``Matrix computations (3rd ed.),'' Johns Hopkins University Press, 1996, USA.

\bibitem{Noughabi:2015}
A.~Noughabi, ``Entropy Estimation Using Numerical Methods,'' in Annals of Data Science, vol. 2, pp. 231--241, Springer Nature, 2015, doi: 10.1007/s40745-015-0045-9.

\bibitem{Han:2023}
C.~S.~Han and K.~M.~Lee, ``Gradient-based Counterfactual Generation for Sparse and Diverse Counterfactual Explanations,'' in Proceedings of the 38th ACM/SIGAPP Symposium on Applied Computing (SAC '23), Association for Computing Machinery, New York, NY, USA, pp. 1240--1247, 2023, doi: 10.1145/3555776.3577737.

\bibitem{Kullback:1951}
S.~Kullback and R.~A.~Leibler, ``On Information and Sufficiency,'' The Annals of Mathematical Statistics, vol. 22, no. 1, 1951, pp. 79–86. 

\bibitem{Eriksson:2004}
K.~Eriksson, D.~Estep, and C.~Johnson, ``Lipschitz Continuity,'' in Applied Mathematics: Body and Soul, Springer, Berlin, Heidelberg, 2004, doi: 10.1007/978-3-662-05796-4\_12

\end{thebibliography}
\end{document}